\documentclass{egpubl}
\usepackage{3dor21}

\WsSubmission      % uncomment for submission to EG Workshop
\usepackage[T1]{fontenc}
\usepackage{dfadobe}  

\usepackage{cite}  % comment out for biblatex with backend=biber
\BibtexOrBiblatex
\electronicVersion
\PrintedOrElectronic
\ifpdf \usepackage[pdftex]{graphicx} \pdfcompresslevel=9
\else \usepackage[dvips]{graphicx} \fi

\usepackage{egweblnk} 
\title[Multi-resolution deep learning pipeline]%
      {Multi-resolution deep learning pipeline for dense large scale point clouds}

\author[Thomas Richard, Florent Dupont \& Guillaume Lavoue]
{\parbox{\textwidth}{\centering Thomas Richard, Florent Dupont and Guillaume Lavoue
        }
        \\
{\parbox{\textwidth}{\centering CNRS, Université de Lyon, LIRIS, France\\
       }
}
}

\begin{document}

\maketitle
%-------------------------------------------------------------------------
\begin{abstract}

Recent development of 3D sensors allows the acquisition of extremely dense 3D point clouds of large-scale scenes.
%Only few deep learning approaches are able to perform semantic segmentation of such large point clouds.
% The main challenge remain in the size of the data, which induce expensive computational and memory cost.
% This issue gets worse when processing full resolution cloud to benefit from all details.
The main challenge of processing such large point clouds remains in the size of the data, which induce expensive computational and memory cost.
%In this context, the full resolution cloud is almost never exploit.
In this context, the full resolution cloud is particularly hard to process, and details it brings are rarely exploited.
Although fine-grained details are important for detection of small objects, they can alter the local geometry of large structural parts and mislead deep learning networks.
In this paper, we introduce a new generic deep learning pipeline to exploit the full precision of large scale point clouds, but only for objects that require details.
The core idea of our approach is to split up the process into multiple sub-networks which operate on different resolutions and with each their specific classes to retrieve. 
% Orig %Thus, the pipeline allow each network to benefit either from noise and memory cost reduction of a sub-sampling or from fine-grained details, according to the object type to retrieve.
% V1 %Thus, the pipeline benefit either from noise and memory cost reduction of a sub-sampling or from fine-grained details, according to the object type to retrieve.
Thus, the pipeline allows each class to benefit either from noise and memory cost reduction of a sub-sampling or from fine-grained details.

%-------------------------------------------------------------------------
%  ACM CCS 1998
%  (see https://www.acm.org/publications/computing-classification-system/1998)
% \begin{classification} % according to https://www.acm.org/publications/computing-classification-system/1998
% \CCScat{Computer Graphics}{I.3.3}{Picture/Image Generation}{Line and curve generation}
% \end{classification}
%-------------------------------------------------------------------------
%  ACM CCS 2012
% (see https://www.acm.org/publications/class-2012)
%The tool at \url{http://dl.acm.org/ccs.cfm} can be used to generate
% CCS codes.
%Example:


\end{abstract}  
%-------------------------------------------------------------------------

\section{Introduction}

Recent development of 3D acquisition technologies presents several new challenges to the semantic segmentation of 3D point clouds.
%In addition of being unstructured, unordered and irregularly sampled, point clouds can now contain very large scale scenes which add new efficiency constraints.
In addition to being unstructured, unordered and irregularly sampled, point clouds can now contain very large scale scenes which induce higher computational and memory cost.
Moreover, 3D point cloud semantic segmentation requires the understanding of both large scale geometric structure and detailed geometry of the scene.
Obtaining these two elements is even harder with the substantial scale difference brought by large scale point clouds.

Few works have succeeded in processing these massive point clouds.
Among them, RandLA-Net \cite{randlanet2020hu} can process up to 1 million points in a single pass with a smart use of random sampling.
SPG \cite{landrieu2018largescale} uses a superpoint graph as an intermediate structure to learn from clouds with several million points.
Flex-Convolution \cite{flexconvolution2020groh} proposes a new convolution kernel designed to benefit from GPU acceleration, which allows very large point cloud processing. 

\begin{figure}[htb]
  \centering
  \includegraphics[width=.49\linewidth]{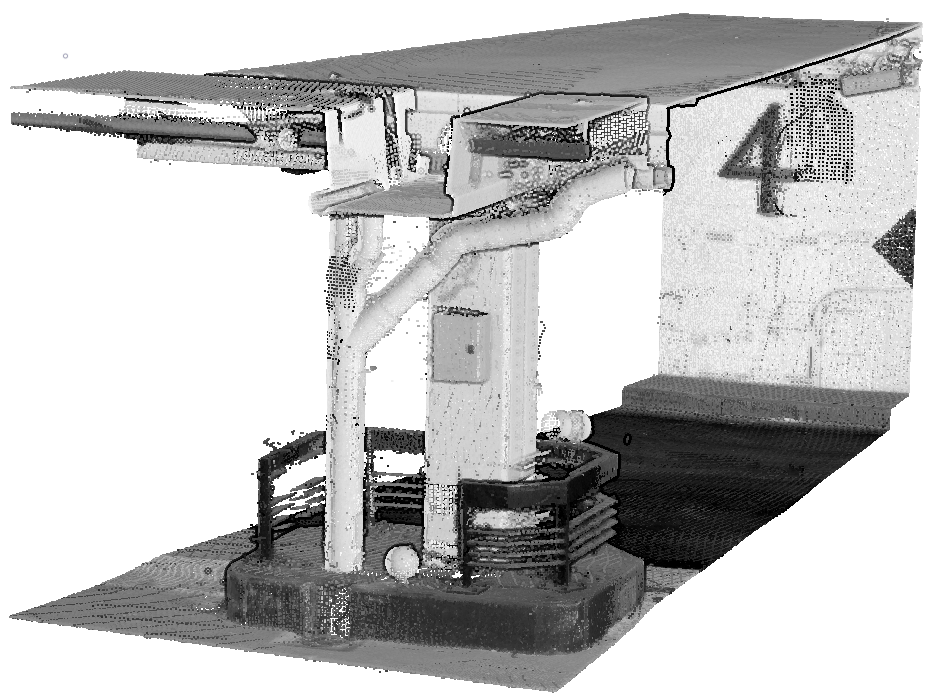}
  \includegraphics[width=.49\linewidth]{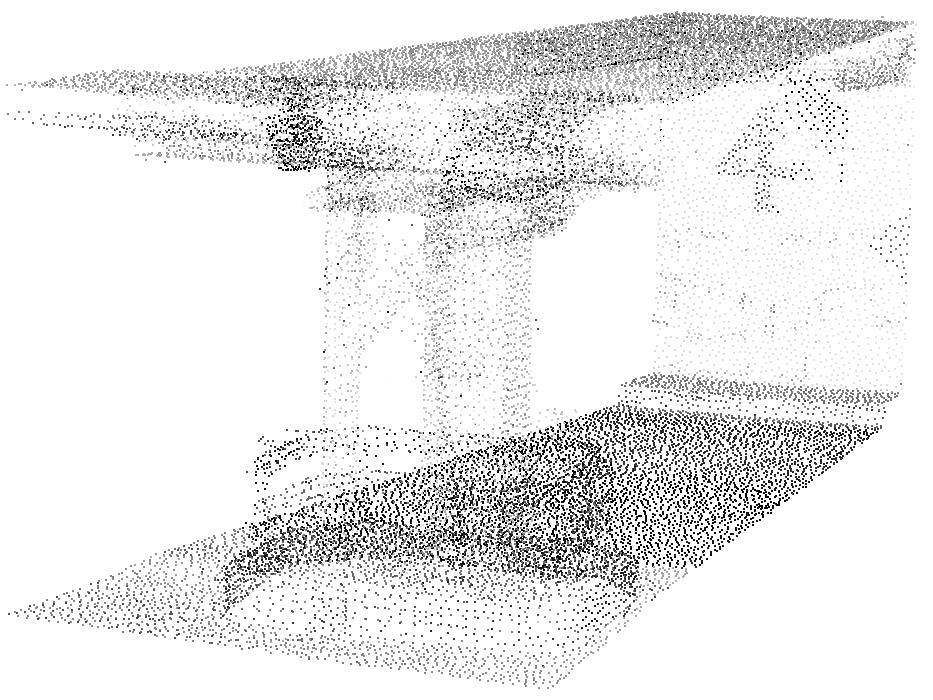}
  
  %\caption{Comparison of different cloud voxelisation. Left: the full resolution cloud. Righ: resolution with one point in each 3cm voxel, which is the subsample size used in the first step of our pipeline.}
  \caption{\label{fig:res} Comparison of different densities of a cloud from LPA dataset. Left: original LPA density, 1 million points per m$^2$. Right: density of the S3DIS dataset, 0.2 million points per m$^2$.}
\end{figure}

All of these approaches have been tested on publicly available dataset like semantic3D \cite{semantic3D2017hackel} or S3DIS \cite{S3DIS2017armeni}.
These datasets contain scenes with each approximately 0.1 and 0.2 million points per m$^2$.
Our work will focus on extremely dense point clouds with up to 1 million points per m$^2$, see figure \ref{fig:res}.
These clouds are provided by the LPA dataset, see section \ref{sec:exp}, a set of 23 labeled 3D point clouds from an underground car park.
It contains large scale objects, like ground or walls, as well as small scale objects, like electric boxes or extinguishers.%, see figure \ref{fig:results}. 

High density point clouds allow to capture much more fine details.
However, to exploit the latter a segmentation needs to operate at full resolution which induces high memory and computational cost.
Although details are useful to segment detailed objects, they can become problematic for large scale objects due to noise or small geometric artefacts that can alter their local geometry.

To tackle these issues we propose a new generic deep learning pipeline which adapts the cloud resolution according to the suitable level of details for the segmentation of each object.
This approach exploits the full cloud precision but only for objects that require details, which allow its usage even on large scale point clouds.
To do so, we split up the segmentation into multiple sub-networks which operate on different resolutions and with each their specific objects to segment. 
Although this approach can be used with any deep learning framework, we used it in combination with the Super-Point graph framework \cite{landrieu2018largescale} for our experiment.

% Old format
% \begin{figure}[t!]
%     \centering
%     \begin{subfigure}[t]{0.25\textwidth}
%         \centering
%          \includegraphics[width=\textwidth]{figures/full_res.png}
%         \caption{Full cloud}
%     \end{subfigure}%
%     \begin{subfigure}[t]{0.25\textwidth}
%         \centering
%          \includegraphics[width=\textwidth]{figures/low_res.png}
%         \caption{3cm}
%     \end{subfigure}
%     %\caption{Comparison of different cloud voxelisation. Left: the full resolution cloud. Righ: resolution with one point in each 3cm voxel, which is the subsample size used in the first step of our pipeline.}
%     \caption{Comparison of different cloud resolution. Left: the full resolution cloud. Righ: cloud with a 3cm voxelisation, which is the resolution used in the first step of our pipeline.}
% \end{figure}
\section{Related work}

In this section we will briefly present different deep learning frameworks designed to tackle the problem of the semantic segmentation of 3D point cloud scenes.

\subsection{Grid based}
As point clouds are unstructured, a natural way to process is to perform a projection into a structured data structure.
%Thus, early approaches propose to embed point clouds into 3D voxel structure and operate convolution using 3D kernels \cite{octnet2017riegler,segcloud2017tchapmi,vnet2016milletari,volumetric2016qi}.
Thus, early approaches propose to embed point clouds into 3D voxel structure and operate convolution using 3D kernels \cite{octnet2017riegler,segcloud2017tchapmi,volumetric2016qi}.

%Other methods use advances of matured 2D CNNs by rendering 3D point clouds into sets of 2D images from different points of view \cite{proj2017lawin,unstructured2017boulch,multiview2017chen,volumetric2016qi}.
Other methods use advances of matured 2D CNNs by rendering 3D point clouds into sets of 2D images from different points of view \cite{unstructured2017boulch,multiview2017chen,volumetric2016qi}.

\subsection{MLP based}

The pioneer work PointNet \cite{qi2017pointnet} directly consumes point cloud by learning pointwise features independently with several shared Multi-Layer Perceptrons (MLPs).
However, this type of architecture cannot capture the relations between points and therefore the local geometry.
To process a wider context, several approaches propose to use information from local neighborhood  \cite{pointsift2018jiang,pointweb2019zhao,shellnet2019zhang,randlanet2020hu}.

\subsection{Convolution based}
Many recent works introduced various designs of convolution kernels for points, which operate directly on point clouds without any intermediate representation \cite{kpconv2019thomas,pointcnn2018li,convpoint2020boulch}.
These approaches rely on the fact that multiple points are needed to form a meaningful shape, and thus perform convolution between points in a local area. 

\subsection{Graph based}
Some approaches design new convolution operators to learn from point clouds represented as a graph structure, in which each point is a node \cite{local2018wang, dynamic2019wang}.
ECC-MV \cite{simonovsky2017dynamic} generalizes the convolution operator to arbitrary graphs of varying size and connectivity. 
GAC \cite{attention2019wang} proposes a Graph Attention Convolution to learn features from a local neighborhood by assigning attention weights. 

\subsection{Large scale based}
Few works focus on segmentation of large scale 3D point clouds.
FCPN \cite{fullyconvolutional2018rethage} uses both voxel and MLP based networks in a fully-convolutional point network able to process clouds with up to 200k points. 
Instead of a more complex point sampling strategy, RandLA-Net \cite{randlanet2020hu} uses a simple but efficient random point sampling, which can process up to 1 million points in a single pass. 
To avoid the potential discard of key features, they introduce a local feature aggregation module to preserve details. 
Flex-Convolution \cite{flexconvolution2020groh} manages to speed up the computation and decrease the memory consumption of convolution based methods with a new convolution kernel defined as a simple scalar product allowing massive GPU acceleration.

The vast majority of the previously presented methods have been designed and evaluated on publicly available dataset like semantic3D \cite{semantic3D2017hackel} or S3DIS \cite{S3DIS2017armeni}.
% Finally, very few methods are able to process large scale point clouds, and 
In contract, our work focuses on the segmentation of large-scale point clouds provided by the LPA dataset, which are much denser.
These are a new type of data to study which opens up new possibilities, especially in the use of fine grained details that are rarely available.
\section{Method}

Our method proposes a generic deep learning pipeline to exploit the full cloud precision only when details are useful to the segmentation.
To do so, we split up the process into multiple sub-networks which operate on different cloud resolutions and with each their optimized learning parameters. 

\subsection{High and low resolution classes}

Low resolution classes are associated with objects that do not need fine details analysis to be segmented.
They generally include large-scale objects like walls, ground or ceiling.
For such classes, details can even bring noise and small unwanted geometry artefacts that can alter their local geometry and thus mislead the network.
%what we call
On the other hand, high resolution classes are associated with detailed objects that can benefit from the precision of a full resolution point cloud.
They generally include small-scale objects such as electrical boxes or mural lights from the LPA dataset.

%Although high resolution labels are generally associated with small-scale objects, and low resolution labels with large-scale objects, there may be exceptions.
However, it is important to point out that the size alone is not sufficient to determine the class resolution.
The details of the local geometry should always be considered.
For example, objects like doors can be seen as large scale objects.
However, they contain fine geometry like handles or frames that help a lot to dissociate them from a wall.
Thus, as they benefit from details, doors are considered as high resolution class.
Another exception are signs from the LPA dataset, they do not benefit from details because of their circular shape that is sufficient to dissociate them from the ceiling, see figure \ref{fig:results}.
They are therefore considered as low resolution class despite their small scale.

\subsection{Multi resolution segmentation}
\label{sec:merge}

To ensure the most discriminating geometry possible, we propose to classify each class at its suitable resolution.
However, in order to adapt the resolution we need to know if each point is considered as high or low resolution class, and an unlabelled cloud does not contain such information.

To overcome this issue, we propose to perform a first segmentation with a different set of classes.
This new set of class is constructed such that all high resolution classes are merged into existing low resolution classes, referred as the concatenated classes, see figure \ref{fig:pipeline}.
Thus we obtain a low resolution cloud populated with low resolution classes only, which are suitable conditions for a segmentation of a low resolution cloud.
Low resolution classes to be merged are chosen according to their adjacency in the scene with the high resolution classes.
As an example, doors and mural lights can be merged into walls, because of their close positioning.
%Moreover all points previously considered as high resolution classes are now labeled as the same low resolution class: the concatenated class.
This first low resolution segmentation with only low resolution classes and the concatenated classes, will then be referred as the initial segmentation.

To retrieve high resolution classes, we simply perform a second segmentation on all points classified as concatenated classes.
As high resolution classes benefit from details, this step is performed at full resolution.
The memory cost is greatly reduced because all large-scale structural objects considered as low resolution classes are not considered.

\subsection{Final results computation}
%For the final result, both low and high resolution segmentation are merged by projecting them on the original full resolution.
The final result clouds are computed by projecting both low and high resolution segmentation results on the original full resolution clouds.

\begin{figure}[htb]
\centering
\includegraphics[width=1.\linewidth]{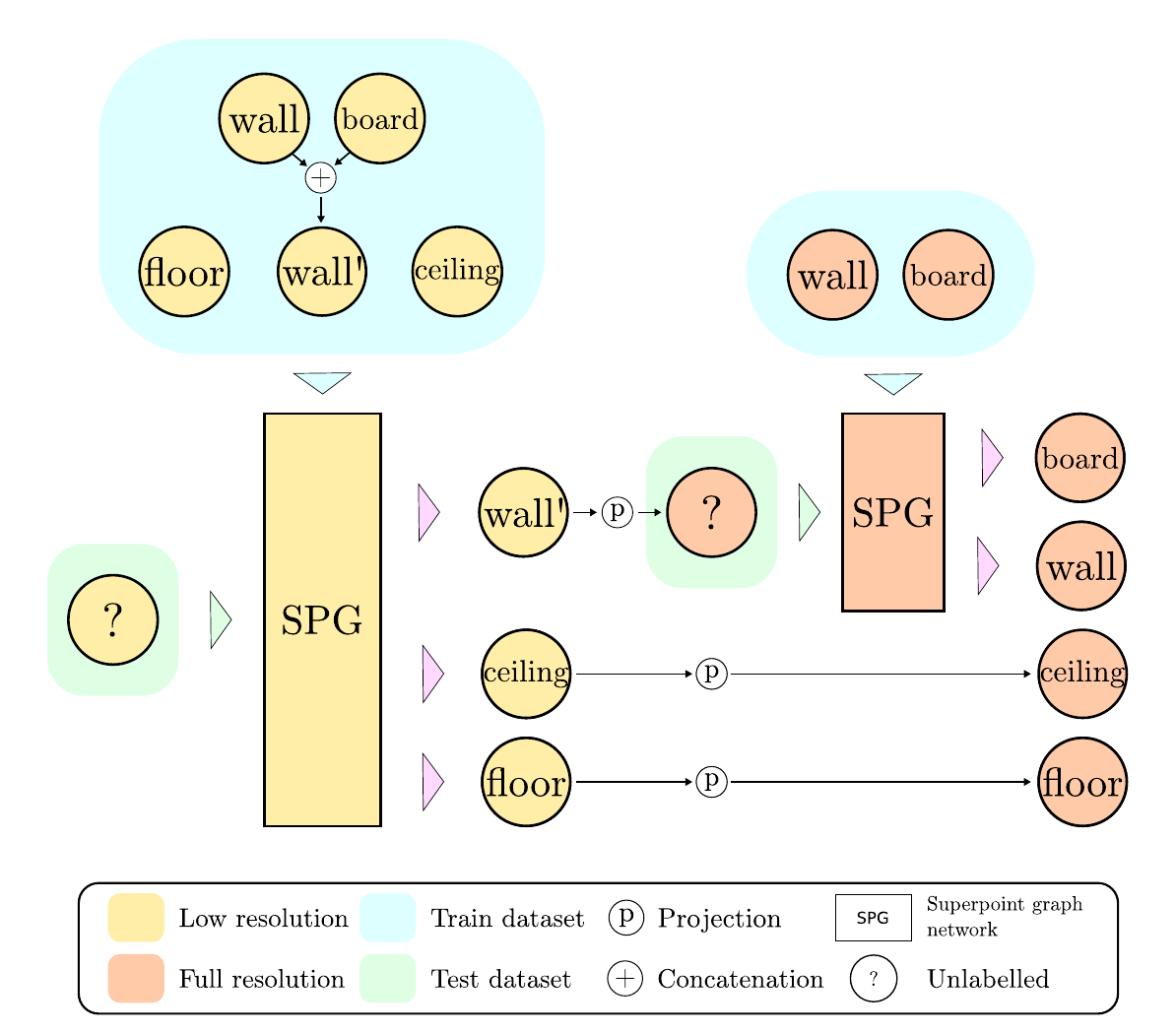}
\caption{\label{fig:pipeline}The proposed multi resolution deep learning pipeline applied to a made-up dataset with 4 classes. In this example $wall'$ is the concatenated class of $wall$ and $board$.}
\end{figure}

The segmented low resolution clouds are projected on the high resolution clouds using a voxel based projection.
%We label all points from the high resolution cloud contained in a voxel with the point label of his voxel counterpart in the low resolution cloud. 
For each voxel in the high resolution cloud, we label all its points with the label of the unique point contained in the corresponding voxel of the low resolution cloud.
The voxel size is the same that was used to subsample the low resolution cloud. 
Finally the segmented high resolution clouds are directly projected on the original high resolution clouds using a closest point projection. 
This operation is necessary as the segmented high resolution clouds have missing parts since they do not contain points associated with low resolution classes, see figure \ref{fig:pipeline}.

%\paragraph{Non coherent semantic}

\section{Superpoint graph}

\label{sec:SPG}

Although our proposed pipeline can be used with any deep learning framework, we used it in combination with the Super-Point graph framework \cite{landrieu2018largescale}.
This section is a brief reminder of this paper work.

\subsection{Geometric partition}

The first step is a weakly supervised over-segmentation of the input cloud into geometrically simple point clusters. 
These clusters are called superpoints.
Points of each superpoint have homogeneous geometric features, therefore it is assumed that they belong to the same object, but without making any assumption about its classification yet.
To better describe the local geometry of each point, 4 features are chosen, proposed by \cite{guinard_weakly_2017}, linearity, planarity, scattering and verticality. 
These features are computed for each point from eigenvalues of the covariance matrix of their respective neighbors.
Superpoints are then modelized using an adjacency graph, as the piecewise constant approximation of a global energy problem.
An approximation of this problem solution is computed using the $l_0$-cut pursuit algorithm proposed by \cite{landrieu:hal-01306779}.

To retrieve entire objects, the relationship between superpoints is modeled by a superpoint graph, in which each node is a superpoint, and edges represent their adjacency relationship.
Each edge has a set of features to bring more information about the relationship, like the centroid offset or surface ratio.

\subsection{Classification}

First, a set of descriptors is computed for each superpoint according to its global shape, by a PointNET \cite{qi2017pointnet} network. 
Points are rescaled to the unit-sphere before their embedding, in order to learn from the superpoint shape and not from its spatial distribution. 
However to stay covariant to the superpoint size, the original metric diameter is concatenated to the final descriptors.

Then to take adjacency between superpoints into account, a contextual classification is performed.
It uses both descriptors previously computed and information from the superpoint graph in a Gated Graph Neural Network (GGNN) \cite{li2017gated}.
Each superpoint is embeded in a GRU initialized with previously computed descriptors from PointNET.
To take edge features into account, the convolution-like operation ECC\cite{simonovsky2017dynamic} idea is used over the superpoint graph.

%OLD

%\subsection{Superpoint embedding}
%
%A set of descriptors is computed for each superpoint according to its global shape, by a PointNET \cite{qi2017pointnet} network. 
%The contextual classification which takes adjacency between superpoints into account is performed as a second step.
%Points are rescaled to the unit-sphere before their embedding, in order to learn from the superpoint shape and not from its spatial distribution. 
%However to stay covariant to the superpoint size, the original metric diameter is concatenated to the final descriptors.
%
%\subsection{Contextual segmentation}
%
%The contextual classification is performed using both descriptors previously computed and information from the superpoint graph using a Gated Graph Neural Network (GGNN) \cite{li2017gated}.
%Each superpoint is embeded in a GRU initialized with previously computed descriptors from PointNET.
%To take edges features into account, the convolution-like operation ECC\cite{simonovsky2017dynamic} idea is used over the superpoint graph.
%\input{method_old}
\section{Experiments}

\label{sec:exp}

\begin{table*}[h]
\begin{center}

\resizebox{1\textwidth}{!}{%
\begin{tabular}{c|cc|cccccccccccc}
  \hline
  & OA & mIoU & Ground & Wall & Ceiling & Sign & Barrier & Box & Fence & Platform & Door & Mur. Light & Elec. Box & Extin.\\
  \hline
  SPG \cite{landrieu2018largescale} & 96.43 & 67.99 & 96.62 & 84.78 & 96.73 & \textbf{87.46} & 79.10 & 72.17 & \textbf{94.32} & 80.28 & 19.70 & 64.79 & 15.94 & 24.02\\
  Ours & \textbf{97.22} & \textbf{76.01} & \textbf{97.88} & \textbf{85.72} & \textbf{96.93} & 84.19 & \textbf{93.68} & \textbf{75.95} & 94.00 & \textbf{88.17} & \textbf{40.73} & \textbf{65.97} & \textbf{40.66} & \textbf{48.26}\\
  \hline
  \hline
  Ours\ (init) & 97.09 & 87.14 & 98.09 & 85.51 & 96.04 & 81.55 & 90.42 & 65.78 & 93.58 & 86.18 & $n/a$ & $n/a$ & $n/a$ & $n/a$\\ 
  \hline
\end{tabular}
}

\end{center}
\caption{Quantitative results comparison on LPA dataset between SPG \cite{landrieu2018largescale} and our deep learning pipeline in combination with SPG. OA is the overall accuracy, the intersection over union is split per class, and mIoU refers to the average of the latter. $init$ is referred as the initial segmentation in our pipeline.}
\label{tab:res}
\end{table*}

%In this section, we evaluate the semantic segmentation of our pipeline on our large-scale indoor dataset named the LPA dataset.

\subsection{Presentation of the LPA dataset}

The LPA dataset is a set of point clouds of an underground car park, which contains 23 clouds from 4 parking floors for a total of 127Mi points.
This dataset is extremely dense and precise, however it contains lots of noise and some minor misalignment issues. 
These properties are explained by its direct origin from the industry, with a minimum cleaning preprocess.
All available classes as well as their association to high or low resolution type are presented in the figure \ref{fig:results}.
As the dataset contains 4 floors, the evaluation will be a 4-folds cross validation.

\subsection{Evaluation}

\begin{figure*}[htb]
    \centering
    \includegraphics[width=1\linewidth]{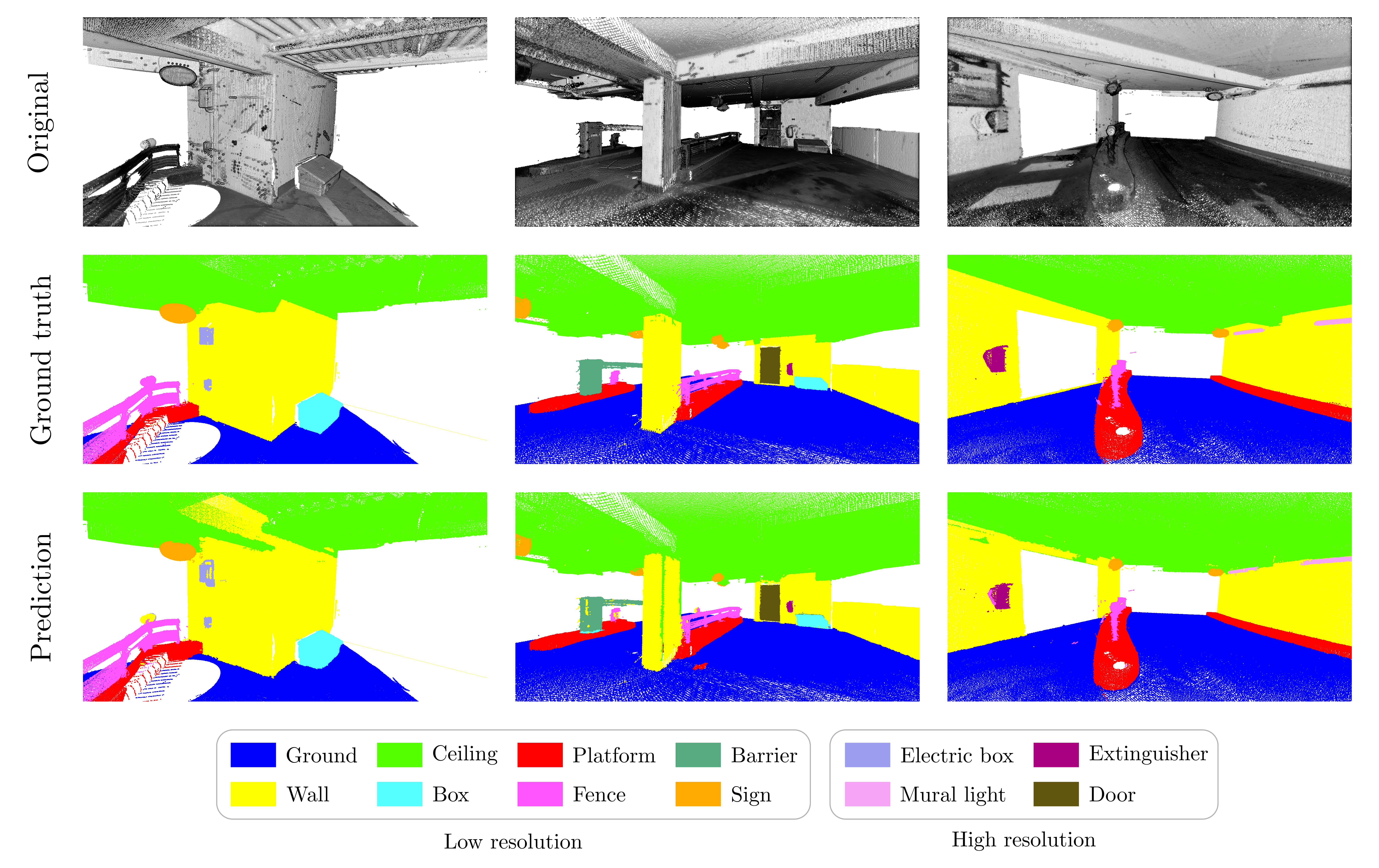}
    \caption{\label{fig:results} Qualitative results on 3 different parts of the LPA dataset, from left to right. From top to bottom: original point cloud, ground truth prediction, predicted classes using our approach.}
\end{figure*}

Results comparison between the original SPG method and our deep learning pipeline in combination with SPG is presented in the table \ref{tab:res}.
Qualitative results of our approach are shown in the figure \ref{fig:results}.
Our method has been evaluated on the full resolution LPA dataset.
For the original SPG, its evaluation at full resolution induces heavy pre-processes that can take several dozen hours per cloud and huge memory consumption.
Therefore we performed a sub-sampling of the clouds before their classification, which is the same strategy used by the original SPG method \cite{landrieu2018largescale}.
The results are then projected using a voxel based projection on the full resolution cloud to obtain comparable results.

We can see that small scale and/or detailed objects like doors, electrical boxes or extinguishers are the hardest classes to retrieve.
It is explained by their particularly detailed geometry and the important point number imbalance they suffer from.
However, fine-grained details brought by our approach lead to substantial improvements in the segmentation of these classes.
The segmentation of large scale objects like ground or walls demonstrate good results for original SPG as well as our approach.
This is the expected behaviour as the segmentation resolution for these classes are identical for both approaches.
We can still observe some minor improvements, especially for classes like barriers of platforms.
They are made possible by the merge of high resolution classes into a single low resolution class.
As the most difficult classes are merged into walls, it limits potential segmentation errors that can confuse contextual information, and thus mislead the method even more.

In table \ref{tab:res}, $init$ is referred to as the initial segmentation in our pipeline, using only low resolution classes including the concatenated class.
We can see differences between our final results and the initial segmentation results, even for low resolution classes.
% These differences can be explained by two factors. 
% First, the step 1 segmentation can contains errors, and some points can be confused as the concatenated class.
% So the second segmentation can change the classification of a point with a ground truth considered as a low resolution class, and thus affect its score.
%%%%%%%%%%%%%%%
% Second, as previously explained a projection of the step 1 results into a high resolution cloud is performed to compute the final results.
%However, as the point cloud density isn't regular, it affect the classes score.
%This is especially true for dense classes.
%In such case a single point correctly classified in the low resolution cloud can represent much more points in the high resolution one.
These differences are induced by the projection of the initial segmentation results into a high resolution cloud, in order to compute the final full resolution results.
Indeed, as the point cloud density is irregular, this operation affects the class's scores.
This is especially true for dense classes, in which a single point correctly classified in the low resolution cloud can represent many more points in the high resolution one.
The exact same projection is used to compute the original SPG results.
We perform a segmentation at low resolution using raw SPG, and then project those results on the high resolution cloud using a voxel based projection, to obtain comparable results.

\section{Conclusion}

In this paper, we presented a new deep learning pipeline to exploit fine-grained details from dense large scale 3D point clouds.
We showed that these details are important to segment certain objects, and introduced new ideas like adaptive density or class merging to process such details on large scale scenes scenario.
This approach leads to better semantic segmentation results of our dataset, composed of dense large scale 3D point clouds.

\section{Acknowledgements}
%This work was performed using XXX resources from XXXXXXXXXXX XXXXXXXXXXXXXXXXXXXXXXXX.
%The LPA dataset was provided thanks to the collaboration of the XXXXXXXXXXXXXX and XXXXXX.
%This work was supported by XXXXXXXXXXXXXXXXXXXX XXXXXX under the XXX XXXXXXX XXXXX XXXXXXX.
This work was performed using HPC resources from GENCI-IDRIS (Grant 2020-AD011012170).
The LPA dataset was provided thanks to the collaboration of the Lyon Parc Auto and Arskan.
This work was supported by Auvergne-Rhône-Alpes region under the R\&D booster grant "CAJUN".

%-------------------------------------------------------------------------

% bibtex
\bibliographystyle{eg-alpha-doi}  
\bibliography{bib}        

% biblatex with biber
%\printbibliography                

\end{document}